# Affordable Modular Autonomous Vehicle Development Platform

Benedict Quartey
*Computer Science Department*
*Ashesi University*
Accra, Ghana
benedict.quartey@alumni.ashesi.edu.gh

G. Ayorkor Korsah
*Computer Science Department*
*Ashesi University*
Accra, Ghana
akorsah @ashesi.edu.gh

***Abstract*—Road accidents are estimated to be the ninth leading cause of death across all age groups globally. 1.25 million people die annually from road accidents and Africa has the highest rate of road fatalities [1]. Research shows that three out of five road accidents are caused by driver-related behavioral factors [2]. Self-driving technology has the potential of saving lives lost to these preventable road accidents. Africa accounts for the majority of road fatalities and as such would benefit immensely from this technology. However, financial constraints prevent viable experimentation and research into self-driving technology in Africa. This paper describes the design of RollE, an affordable modular autonomous vehicle development platform. It is capable of driving via remote control for data collection and also capable of autonomous driving using a convolutional neural network. This system is aimed at providing students and researchers with an affordable autonomous vehicle to develop and test self-driving car technology.**

## I. Introduction

Road accidents are estimated to be the ninth leading cause of death across all age groups globally. The annual estimated global tally of deaths as a result of road accidents hovers around 1.25 million people [1]. These accidents are mostly due to preventable human driver error [2] and autonomous vehicles provide a prospective solution to this problem. Interest in the potential of autonomous vehicles has grown substantially in the past four years. As of June 2017, the research institution Brookings estimated the total investment in research and development of autonomous vehicles by industry leaders to have grown from under $1 billion in late 2014 to about $80 billion [3].

Africa lags behind in this field as the investment data provided by Brookings reveals that the $80 billion research and development transactions and acquisitions stated earlier are situated in already developed economies. While these investments show the interest in autonomous vehicles and spell out an exciting future for artificial intelligence, it also shows the unequal global distribution of knowledge and resources in this field.

African universities and corporations are yet to make attempts to bridge this gap in research into self-driving cars. However, about 90% of the global death toll due to road accidents occur in low and middle-income countries; a category in which most African countries fall. In 2015 alone, 256,719 lives were lost to road accidents on the continent [1]. This single statistic shows the importance of Africa developing capability in self-driving research and development.

Financial constraints are a major roadblock to self-driving research in Africa as typical autonomous vehicle projects require considerable resources. For instance, Stanford university's autonomous vehicle Stanley required dedicated funding from sponsors such as VW, Redbull and Intel, among others [4]. Equipment typically used in these autonomous cars such as Velodyne scanners and ring laser gyroscopes can cost as much as $85,000 and $20,000 respectively [5].

There exists a plethora of commercial and open source robot programming simulation software such as Webots [6], ROS [7] and Gazebo [8] designed to abstract hardware and provide sophisticated tools for single and multi-robot programming. These options seem viable for low cost experimentation involving autonomous robots, and, in fact, a majority of algorithms can be tested in simulated environments. However, the practical usefulness of these simulated environments in validating algorithms for real life probabilistic tasks can be diminished due to the uncertainties, constraints and challenges introduced by the real world [9]. This makes it imperative that self-driving research be conducted on physical hardware subject to these constraints and challenges and necessitates the existence of an affordable physical platform for cost efficient self-driving research.

This paper introduces and describes the design of RollE, a novel affordable modular scaled-down autonomous vehicle platform designed to reduce the barrier to entry into self-driving research, in terms of cost of equipment. This platform aims to accelerate self-driving research in Africa by providing students and researchers with a low cost autonomous vehicle to test ideas and build self-driving technology using machine learning techniques similar to those used in the industry. Fig. 1 shows an image of the RollE Rover (autonomous vehicle).

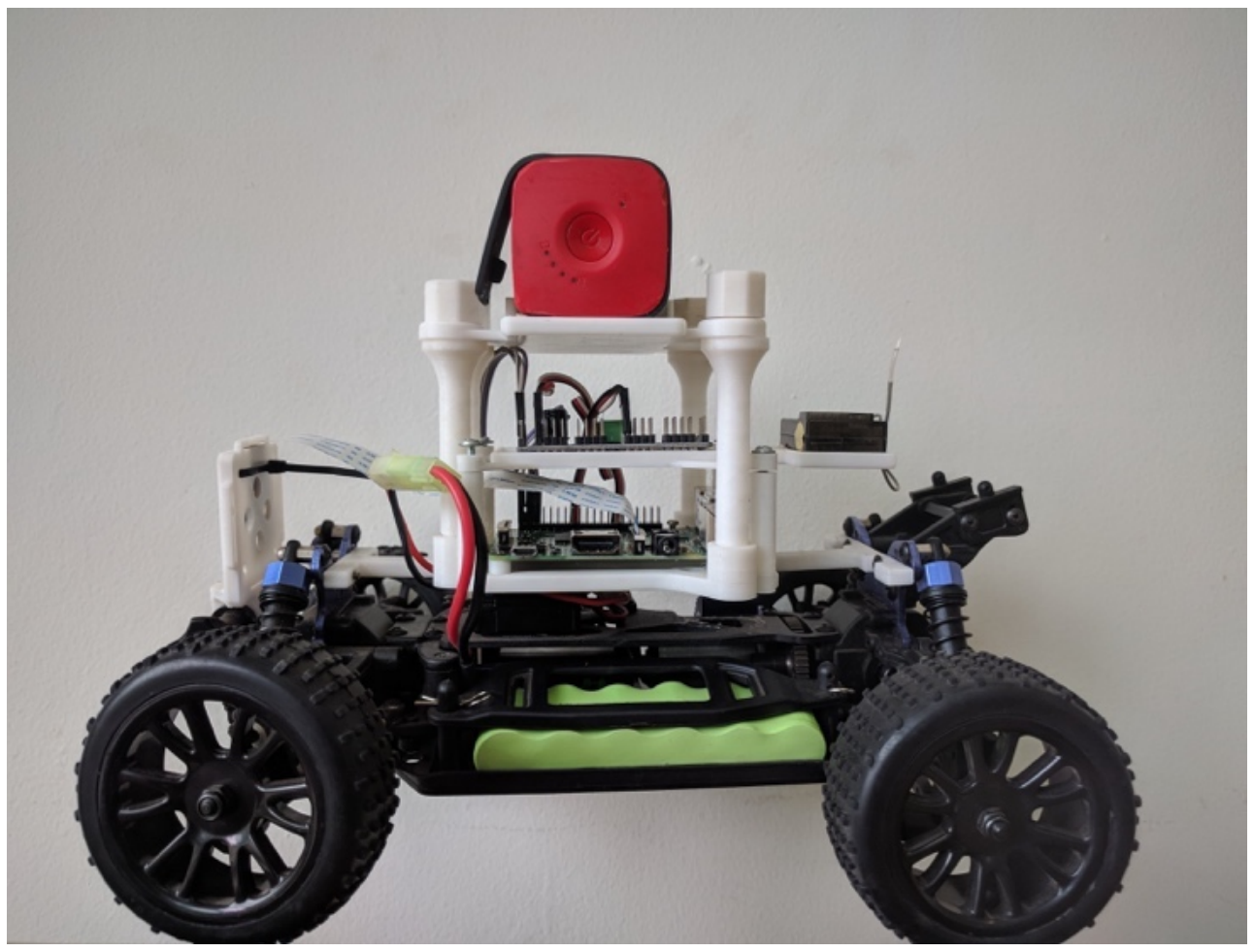

Fig. 1. RollE Rover, a modular autonomous vehicle

## II. Related Work

Progressive developments in the pursuit of self-driving cars has led to technologies such as cruise control and Advanced Driver Assistance Systems (ADAS). These systems have been aimed at extending the sensory capabilities of human drivers to make the driving experience safer. This additional intelligent functionality however is described as level 1 autonomy. RollE attempts level 4 autonomy, this would allow it to act without human input in constrained or specific environments. Breakthroughs in the fields of computer vision, visual convolutional neural networks for image recognition and classification, as well as general advances in machine learning have made achieving such an autonomous system possible.

The most important implementations of autonomous vehicles this project closely models are Nvidia's end-to-end self-driving car experiment [10] and Stanford University's Stanley [11]–winner of the 2005 DARPA Grand Challenge [12].

In Nvidia's implementation of an end-to-end self-driving system, a single front facing camera was used to feed real-time images into a convolutional neural network [10]. This neural network then predicted suitable steering angles. This system used supervised training, thus data from human drivers was used to train the neural network. This particular end-to-end approach to self-driving is what RollE employs.

Stanford University's self-driving car, Stanley applied a more complex approach to self-driving. Unlike Nvidia's implementation that used a single camera, Stanley employed the use of multiple sensor systems for environment perception. These included radars, a camera, laser range finders and GPS antennas. This plethora of sensory data is integrated using an unscented Kalman filter to provide a much more accurate localization system [11]. Research enabling this approach of complex sensor fusion can typically span decades [5]. In the spirit of creating a basic and affordable but scalable and extensible autonomous vehicle development platform, RollE does not use this approach. However, techniques such as decentralized execution of software modules, timestamped data and publish-subscribe based inter-process communication used by the Stanford team improved RollE's end-to-end implementation.

Similar low-cost robotics platforms such as the Evobot [9], ExaBot [13], and Duckiebot [14] have been developed for educative and research purposes. Duckiebot particularly targets self-driving research and provides interesting functionality such as lane following, localization and obstacle avoidance [14]. It also comes in a range of three different configurations based on hardware specifications and functionality.

These systems have proved very useful in the proliferation of robotics research. However, they are based on differential steering and as such do not adequately represent the mechanics of actual wheeled vehicles, which use variations of proportional steering systems. RollE provides an all-in-one proportional steering autonomous vehicle development platform complete with a programmable remote controller for manual wireless driving and data collection. Thus, it provides a platform that more closely represents full-scale autonomous vehicles.

## III. Overview of the RollE Self-Driving System

The machine learning approach this project uses is an end-to-end one, meaning that sensor data–in this case, camera images–is directly passed as input to a machine learning algorithm that would in turn output steering and throttle commands. With this approach, the machine learning model is not explicitly taught to identify hand engineered features such as outlines of roads. The system learns necessary representational features on its own directly from the training data provided.

A front-facing camera connected to the on-board computer (a Raspberry Pi) acts as the single exteroceptive sensory node of RollE. RollE has two primary modes of operation:

- Data collection mode
- Autonomous mode

When in the data collection mode, RollE is controlled by a human agent using either the RollE Pilot (remote controller) or a console remote-control application (Soft Pilot). In this mode, image frames are captured from the camera at a resolution of 200x66 and a frame rate of 32 fps. Each frame is stored with a timestamp and the corresponding throttle and steering values sent from the remote controller at the time of capture. At the end of a data collection run, the images are stored in a folder and the records of steering and throttle commands compiled and saved in a csv file. The data from a data collection run is used to train an end-to-end convolutional neural network based on the architecture used by Nvidia in their self-driving car experiment [10] and implemented using Keras, a neural network application programming interface.

In autonomous mode, RollE is controlled by an autopilot. The camera repeatedly captures frames of its environment and the autopilot software, running locally on RollE, uses the trained convolutional neural network model to predict steering angles for each frame. The throttle value for speed control is set to a constant value. The captured image frames and steering predictions are also transmitted via a socket connection to a user's computer for visualization.

## IV. Design and Implementation of RollE

RollE is an autonomous vehicle built using a 1/16 scale remote control (RC) vehicle as the mobile base. Despite the fact that RollE is built on top of an RC car, similar technologies and machine learning frameworks used in full scale autonomous cars are used in developing this platform. This makes the software system potentially scalable to larger car platforms with minimal changes to the code base. The RollE system can be considered as having four layers: the physical layer which consists of hardware components, and a collection of three logical layers namely Perception, Control and Learning. The term logical layer as used here refers to conceptual layers representing the organization of related software units.

### A. Physical Layer

One of the objectives of RollE is to be modular; as such the system consists of a number of interacting physical components. The hardware components that make up the physical layer are:

- RC Car: this is a 1/16 scale Exceed Blaze RC car that acts as the mobile base of RollE. The Exceed Blaze was chosen for its build quality and proportional steering, which unlike the differential steering mechanism used in some mobile robots [9,13,14], closely imitates the steering mechanism of an actual car. The RC car has actuators and effectors that enable RollE to move in its environment. The Exceed Blaze has the following electronic components:
    - 7.2volt brushed direct current (DC) motor which acts as the throttle motor
    - MG996R servo motor which controls steering
    - WP-1040-Brushed Electronic Speed Controller (ESC)
    - 6 cell 1100 milliamp hour (mAH) 7.2 volt Nickel Metal Hydride (Ni-MH) battery pack
    - 4-channel 2.4 gigahertz (GHz) receiver and transmitter
- RollE Pilot: this is a remote controller designed to enable a user to manually drive the vehicle. It consists of the following components:
    - One Arduino Uno, an open source microcontroller designed to enable reading sensor inputs, performing computation and effecting physical outputs through electronic components. It serves as the brain of the remote controller
    - A pair of XY axes joystick modules, each built from two potentiometers set up in a 2 - dimensional fashion, enabling the movement of a central arm along the X and Y axes to be measured
    - One liquid crystal display (LCD) module which displays information
- Raspberry Pi: this is a single board computer that acts as the on-board processor of RollE. The Raspberry Pi was chosen for its affordable price point, as well its plethora of I/O ports which gives the end users of RollE the freedom to add on various other sensors. The inbuilt general-purpose input/output (GPIO) pins provide an interface to programmatically communicate with low-level sensors and electronic devices such as the ESC and servo motor on the Exceed Blaze
- Raspberry Pi Camera: this is a single board module fitted with a 5MP Omnivison 5647 fixed focus sensor. It is capable of taking high resolution images and gives RollE the ability to perceive its environment
- Router: this creates an on-board wireless network that enables a user to wirelessly communicate with RollE over considerable distances
- Adafruit PCA9685 16-channel pulse width modulation (PWM) driver: this component enables programmatic control of the steering servo motor and throttle DC motor on the Exceed Blaze via pulse width modulation signals

These hardware components are assembled to form the complete RollE system which consists of the RollE Rover (mobile vehicle) and RollE Pilot (remote controller). The cost of the complete RollE system is under $250.

***RollE Rover:*** The default setup of the RC car wired the battery pack and throttle motor to the electronic speed control. The steering servo and ESC were connected to separate channels on the 2.4 gigahertz receiver, enabling the car to be controlled via a radio transmitter. We maintained the connection between the battery pack, ESC and throttle motor. However, the steering servo and ESC were rewired from the receiver to individual channels on an Adafruit PCA9685 PWM driver. The PWM driver was then connected to a Raspberry Pi via the on-board ground (GND), voltage common collector (VCC), serial clock (SCL) and serial data signal (SDA) GPIO pins. This connection enables programmatic control of the servo and throttle motors via pulse width modulation. The technique of pulse width modulation enables a user to programmatically vary the speed of the throttle motor or specify the desired angle of turn of the steering servo motor. A Raspberry Pi camera, which acts as RollE's visual sensory node, was also connected to the Raspberry Pi. Fig. 2 illustrates the final component connections for the RollE Rover.

A collection of modular 3D printable components designed in Autodesk Fusion 360 provide a functional and aesthetic way to attach all the components of the RollE Rover to the mobile RC car platform. These components are designed to vertically stack on top of a base structure. This enables end users to vertically stack additional components and sensors.

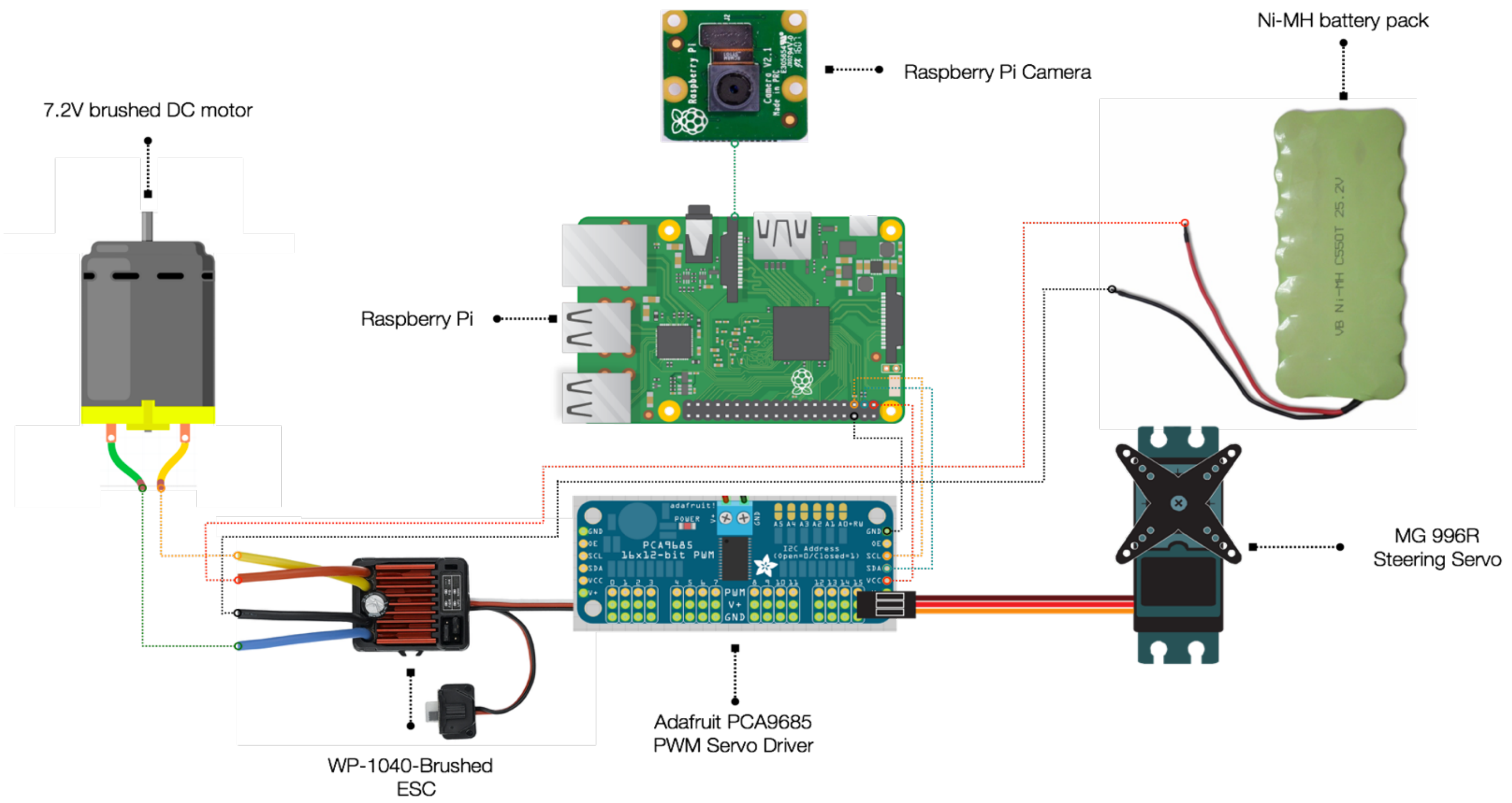

Fig. 2. Diagram of the RollE Rover's component connections

***RollE Pilot:*** An Arduino Uno acts as the brain of the remote controller; it takes sensor readings, performs computation and outputs information. A pair of XY axes joysticks acts as the sensors in this system; each joystick is connected to the Arduino in a fashion that restricts readings to one axis per joystick. With this configuration, one joystick takes readings from the X-axis and controls throttle while the other takes readings from the Y-axis and controls steering. An LCD module displays the real-time readings from each joystick and prints out error messages. Sensor readings from joysticks are analog and fall in a range of values from 0 to 1023. These values are remapped to a range between -1 to 1 and are transmitted via serial communication from the Arduino to a python script that separates each joystick's reading and broadcasts the values to the RollE Rover for actuation. Details on this communication between the remote controller and the rover are discussed in the logical layers section, under the control layer.

Much like the RollE Rover, the remote controller also has a collection of 3D printable components designed to hold all components in a functional and aesthetic fashion. The base 3D component of the remote controller allows it to be attached to a standard breadboard, enabling end users to experiment with the remote and attach additional components. Fig. 3 shows a diagram of the component connections for the RollE pilot.

## B. *Logical Layers*

The logical layers of the system each contain a collection of related software units or processes that work together to accomplish some goal. Processes within a layer can communicate with each other as well as communicate with processes from other layers. Inter-process communication is achieved through a publish/subscribe-based communication system. This allows processes to subscribe to specific topics of interest in order to receive broadcast messages published on those topics by other processes. RollE's communication architecture is implemented using the lightweight Message Queuing Telemetry Transport (MQTT) protocol built for connecting devices on networks with minimal bandwidth. RollE consists of the following interacting logical layers:

- *Control:* the software units in this layer serve as an interface between the user and the hardware components of the physical layer. They are also responsible for converting steering predictions from processes in the learning layer into specific pulse values the PCA9685 PWM module can understand. This layer also contains a command-line software implementation of the RollE Pilot (remote controller). This console application gives users discrete control of the throttle and steering values that drives RollE.

- *Perception:* the software units in this layer mainly implement computer vision procedures for capturing, formatting and either transmitting or locally storing images obtained from the on-board camera sensor.

- *Learning:* the software units in this layer deal with machine learning. They specify the architecture for machine learning models and contain code that manage data processing, data augmentation, training the machine learning model and transmitting predictions from trained models to other processes. The machine learning model architecture of choice for this problem domain is a convolutional neural network. Convolutional neural networks are multilayer perceptron machine learning algorithms optimised for analysing visual data and feature extraction by incrementally applying convolutional operations to images at certain layers of the network.

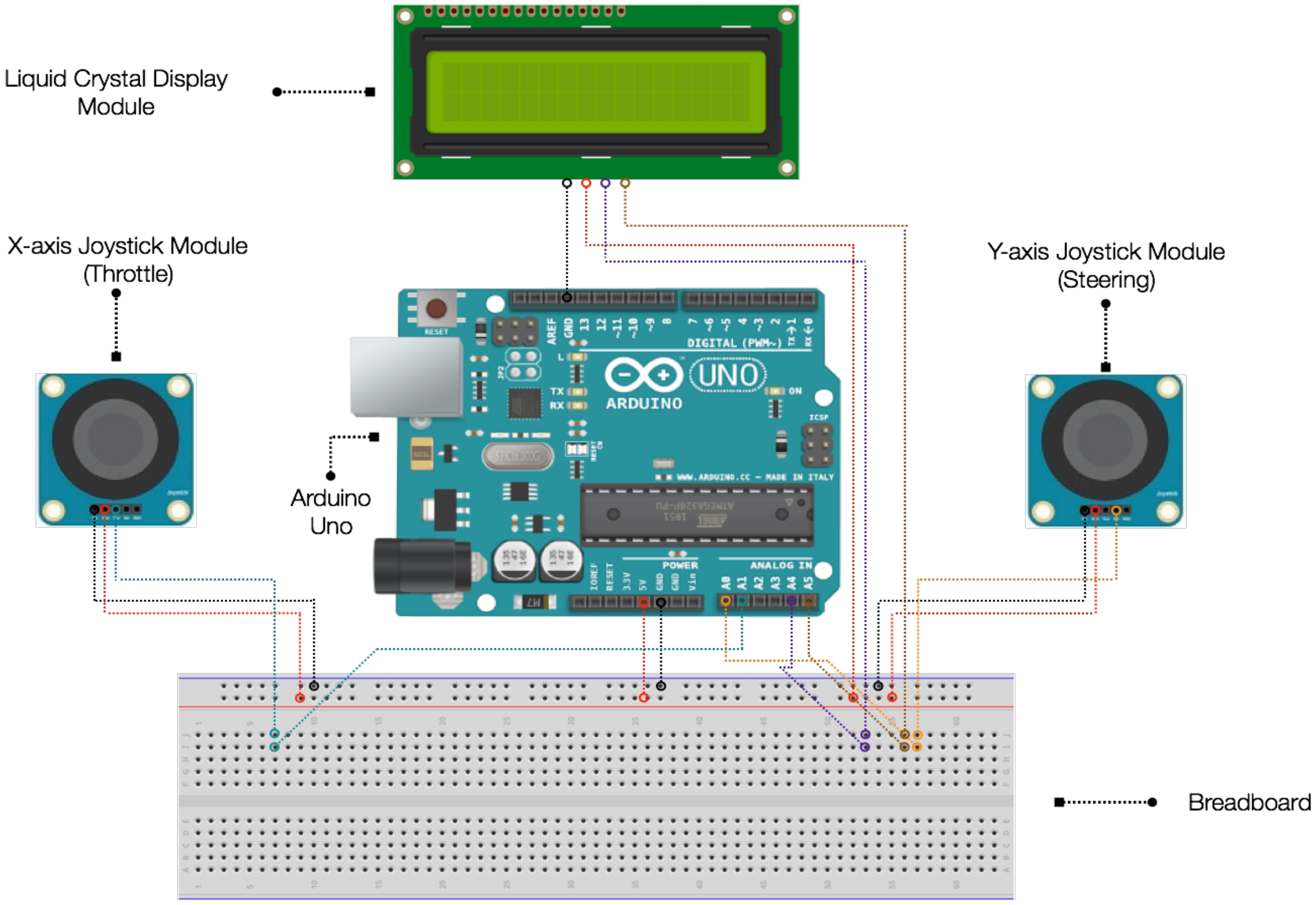

Fig. 3. Diagram of RollE Pilot's component connection

*1) Control Layer*

This layer contains a collection of software units that directly interface with the components of the physical layer. The software in this layer has three primary functions:

- provide programmatic control of the steering and throttle motors of the RollE rover
- take joystick position readings and control the LCD display on RollE Pilot
- transmit steering and throttle values from RollE Pilot to the rover

Four software units “Actuation.py”, “RollE_Pilot.ino” and “Pilot_Transmitter.py / Soft_Pilot.py” respectively provide this functionality.

*Actuation.py:* This python script runs on the Raspberry Pi on the RollE rover and uses the Adafruit_PCA9685 library to specify which signals the PWM module sends to steering and throttle motors. Input values are remapped from a range of -1 to 1 into pulse signals that correspond to maximum and minimum duty cycles. This technique associates the maximum and minimum steering servo angles and throttle motor speeds with a value of -1 and 1 respectively. All values between this range then represents different steering positions or throttle speeds. As an example, a steering value of -1 would turn the steering servo to its maximum position on the left, while a steering value of 0 would turn the steering servo to its middle position; similar logic applies to throttle speeds.

Limiting motor control inputs to the range -1 to 1 simplifies controlling RollE for end users as they can create higher level programs and only have to think of steering and throttle outputs in terms of proportionality as opposed to raw duty cycles. This also allows users to make simple modifications to global maximum and minimum duty cycle values in Actuation.py to change the range of responsiveness or sharpness of steering turns without changing the outputs from their higher-level programs. An example of such higher-level programs is the software implementation of the RollE pilot. This program takes keyboard inputs from users and outputs steering and throttle values in the range -1 to 1 to control RollE.

*RollE_Pilot.ino:* This Arduino sketch controls the RollE Pilot. It takes readings from the 2-dimensional potentiometers that make up each joystick module and remaps these analog values from a range of 0 to 1024 to values in the range -1 to 1. It also controls a 20x4 LCD module which displays the real-time remapped steering and throttle values from the joysticks. This code is flashed and ran on the onboard microcontroller of the RollE Pilot.

*Pilot_Transmitter.py / Soft_Pilot.py:* The pilot_Transmitter.py script communicates with the RollE Pilot via a serial interface and transmits steering and throttle values obtained from the controller to the RollE rover via the MQTT protocol. Soft_Pilot.py is also a command-line software implementation of the physical remote controller; it accepts user keyboard inputs and outputs steering and throttle values in the -1 to 1 range to RollE also via the MQTT protocol. A user connects the RollE Pilot to their computer and the Pilot_Transmitter.py program, which runs on the user’s computer, transmits real-time steering and throttle values to the “RollE_MKII/throttle” and “RollE_MKII/steering topics” respectively. Actuation.py subscribes to these topics, thus receives the transmitted steering and throttle values and converts them to PWM duty cycles to drive RollE.

*2) Perception Layer*

The software units in this layer implement computer vision procedures for capturing, formatting and either transmitting or locally storing images obtained from the on-board camera sensor of the RollE rover. In the data collection mode, images are stored locally on the Raspberry Pi. In autonomous mode, images are served as inputs to a trained neural network model running locally on the Raspberry Pi for steering predictions. These images can also be transmitted via a socket connection to the user's computer for visualization. The computer vision pre-processing operations performed on each image frame as shown in Fig. 4 are:

i. a 180-degree rotation on captured images, done because restrictions in the length of ribbon cable that connects the Pi camera to the Raspberry Pi forced the camera to be placed upside down during assembly of the rover

ii. image is cropped to remove unwanted parts of image above track.

iii. image is resized to fit the shape that the convolutional neural network accepts. The acceptable shape must have an image height of 66 pixels, width of 200 pixels and 3 colour channels (3@66x200)

iv. the colour model of the image frame is converted from RGB to the YUV colour space. The YUV colour space represents images with one luma component (Y) and two chrominance components (UV). The input image is split into these individual YUV planes before being passed into the convolutional network for feature extraction.

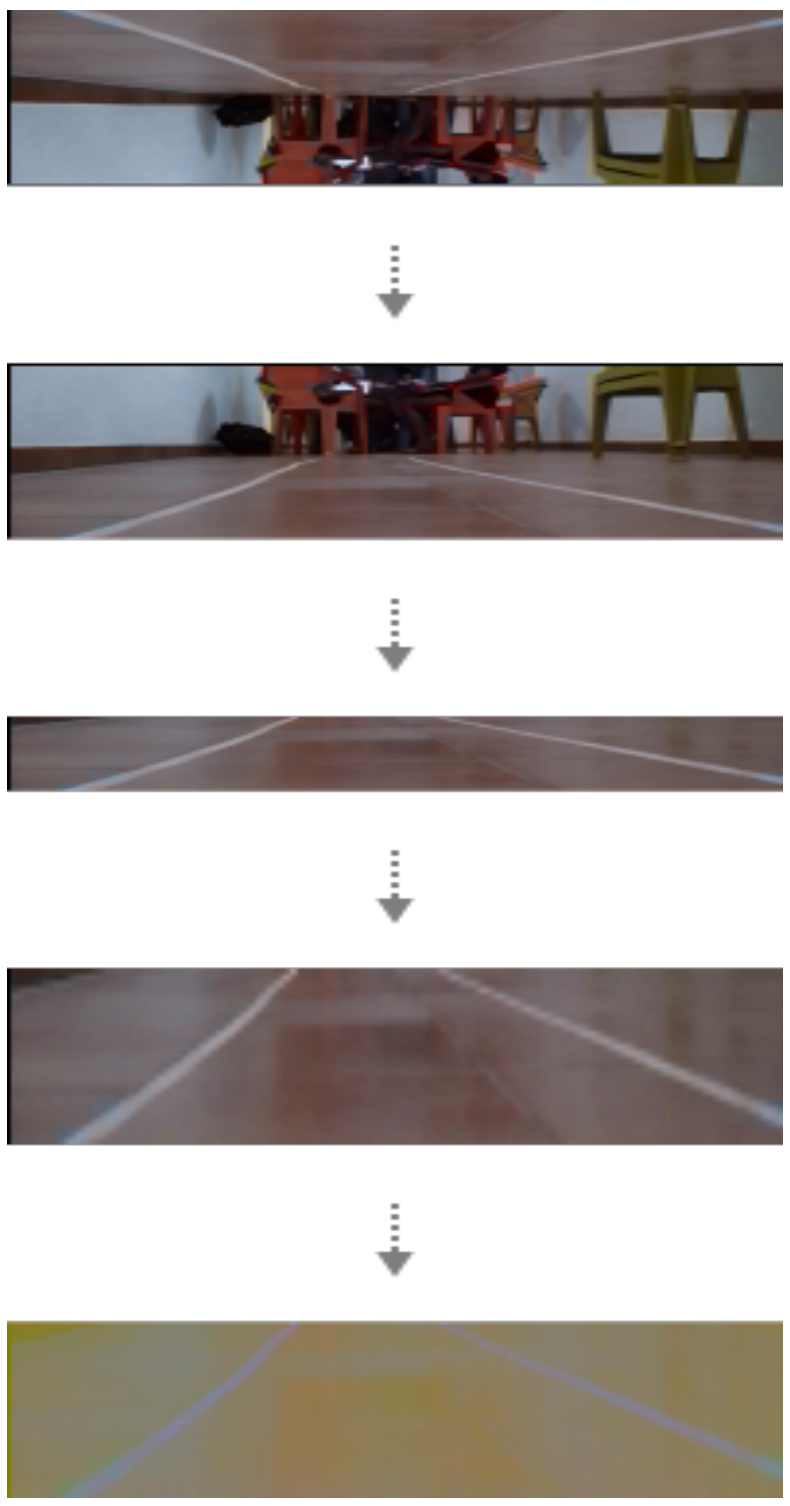

Fig. 4. Sequence of image pre-processing steps. First image in sequence is an original upside-down image as captured by RollE

*3) Learning Layer*

The software units in this layer deal with machine learning. They specify the architecture of machine learning models and contain code that manage data processing, data augmentation, training the model and transmitting predictions from trained models to other processes. The model used for this system is a convolutional neural network based on the end-to-end architecture shown in Fig. 5 and used by Nvidia in their self-driving car experiment. For this project this convolutional neural network architecture was implemented using Keras, a deep neural network application programming interface–with Tensorflow as the backend.

After data is collected during a data collection run, the recorded collection of image frames and steering labels are used for training. Data augmentation remains one of the easiest methods to reduce overfitting in convolutional neural networks [15], as such it is performed on the training examples to produce variations that help the model learn more from existing examples and reduce overfitting. Two key data augmentation operations, random horizontal flips and random shadowing are performed on training examples. Randomly selected training examples are horizontally flipped, and their corresponding steering angles negated to create new data samples. Random shadows are also created on some training examples to help the model learn how to handle real-world scenarios where sections of captured images are dark due to shadows from objects. Much like the approach used by Krizhevsky et al. in AlexNet [15], the augmentation is done in real time during training by a python batch generator which allows generation of images without storing to disk. Fig. 6 shows sample augmented images.

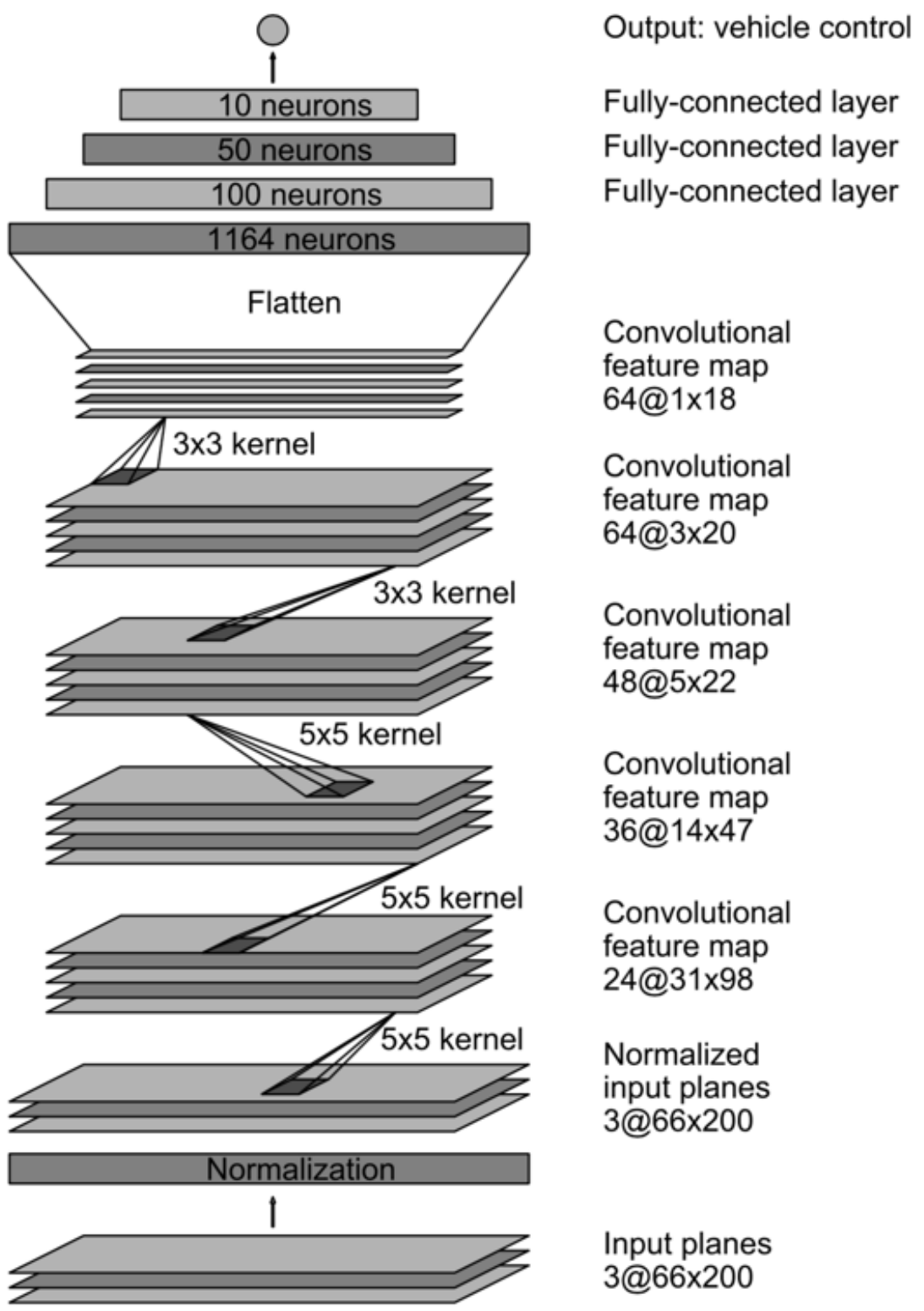

Fig. 5. End to End convolutional neural network architecture developed by Nvidia researchers [11]

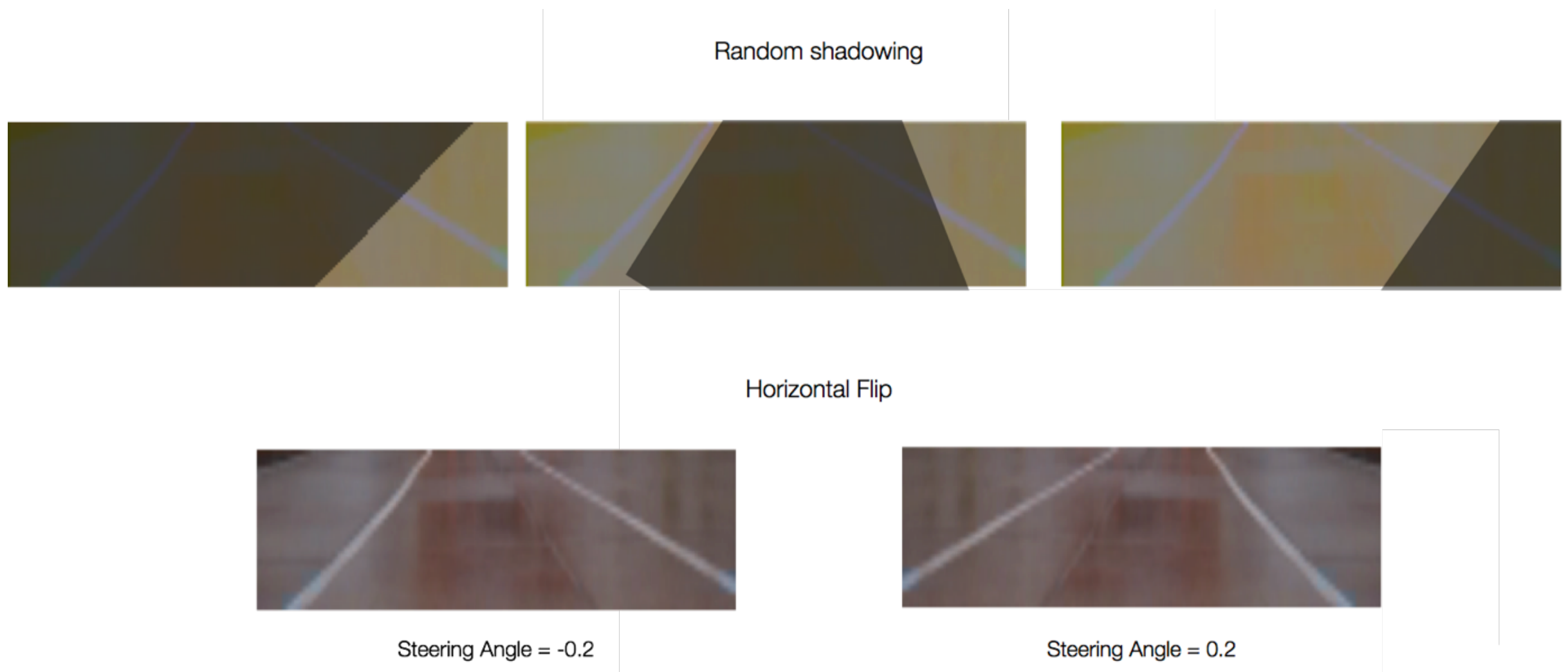

Fig. 6. Sample images showing results of data augmentation

## V. Evaluation

Two tests were conducted to evaluate the two primary modes of RollE: the data collection mode and the autonomous mode.

### A. *Data collection*

To ensure that the entire pipeline of data collection processes functioned optimally, we drove the RollE Rover with the remote controller in data collection mode on a tiled path bordered on both sides by lawns as shown in Fig. 7. At the end of this run, a total of 3147 images were captured and stored with timestamps on the processor of the rover. A comma separated values (CSV) file with the paths to the stored images labelled with steering and throttle values was also created at the end of the data collection run. This confirmed that the data collection process worked as expected.

### B. *Autonomous drive*

The autonomous drive test ensured that once a model has been trained, RollE can successfully drive itself on tracks similar to what it was trained on. The convolutional neural network model was trained using the data collected in the data collection test described earlier. Eighty percent of the gathered data was used as the training set and the remaining twenty percent as the validation set. The model was trained for 10 epochs with a learning rate of 0.0001. Performing the random data augmentation operations described earlier, we generated and trained on 20,000 samples each epoch. The mean squared error of the trained model on the validation set at the final epoch was an impressive 0.028. Fig. 8 shows a graph of the mean squared errors of the model plotted for both validation and training set at each epoch.

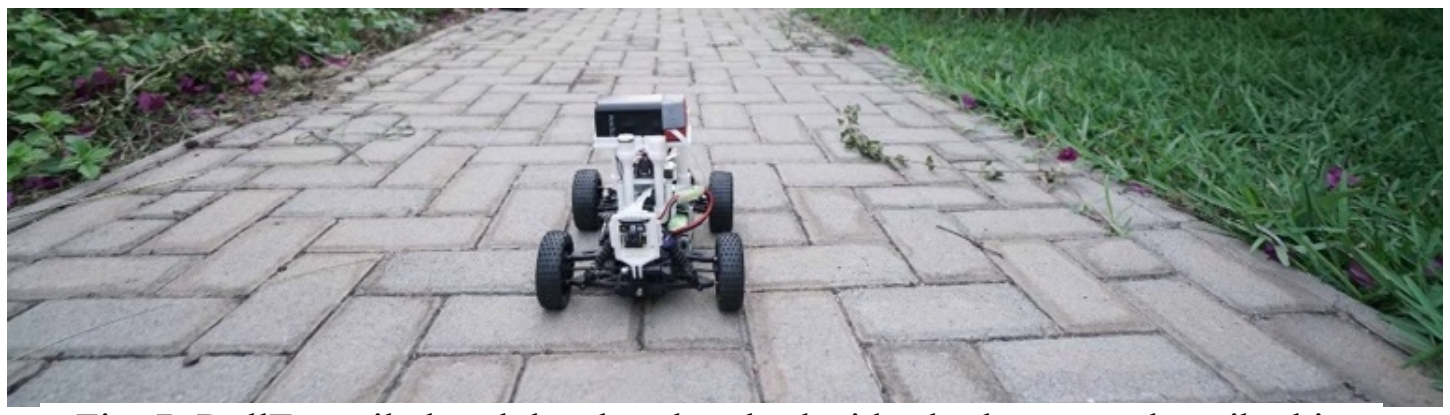

Fig. 7. RollE on tiled path bordered on both sides by lawns as described in Data collection section

After training the model, RollE was set on the tiled path and put in autonomous mode. It was observed that RollE exhibited intelligence and intentionality in its driving. It successfully made corrective turns to avoid the lawns and stay on the tiled path. However, RollE seemed to favour driving towards the left and after a couple of meters, it completely swayed to the left and ended up on the lawn. This behaviour was due to the distribution of the training data; while collecting data we favoured steering towards the left on the remote controller. The horizontal flipping data augmentation operation helped balance out this data. However, the magnitude of training examples with steering towards the left forced this behaviour to be incorporated in the model.

Fig. 9 shows the distribution of steering values for a random sample of 100 training examples. It can be seen that, most of the steering was either towards the left (-1) or in the resting position (0) in the original data. After performing data augmentation, the data distribution became more balanced. However, it is clear that left steering values would still be favoured by the predictive model. The limitation of data depending on the ability of a human agent necessitates that the human agent generating the data drives RollE in a fairly consistent fashion. Future work could experiment with some self-supervised learning techniques as proposed by Stavens [5], to eliminate the possibility of inconsistent data due to human error. RollE was driven once again in a fairly consistent manner on a different track shown in Fig. 10. A video showing a data collection and autonomous driving demonstration of RollE on this track can be seen in [16].

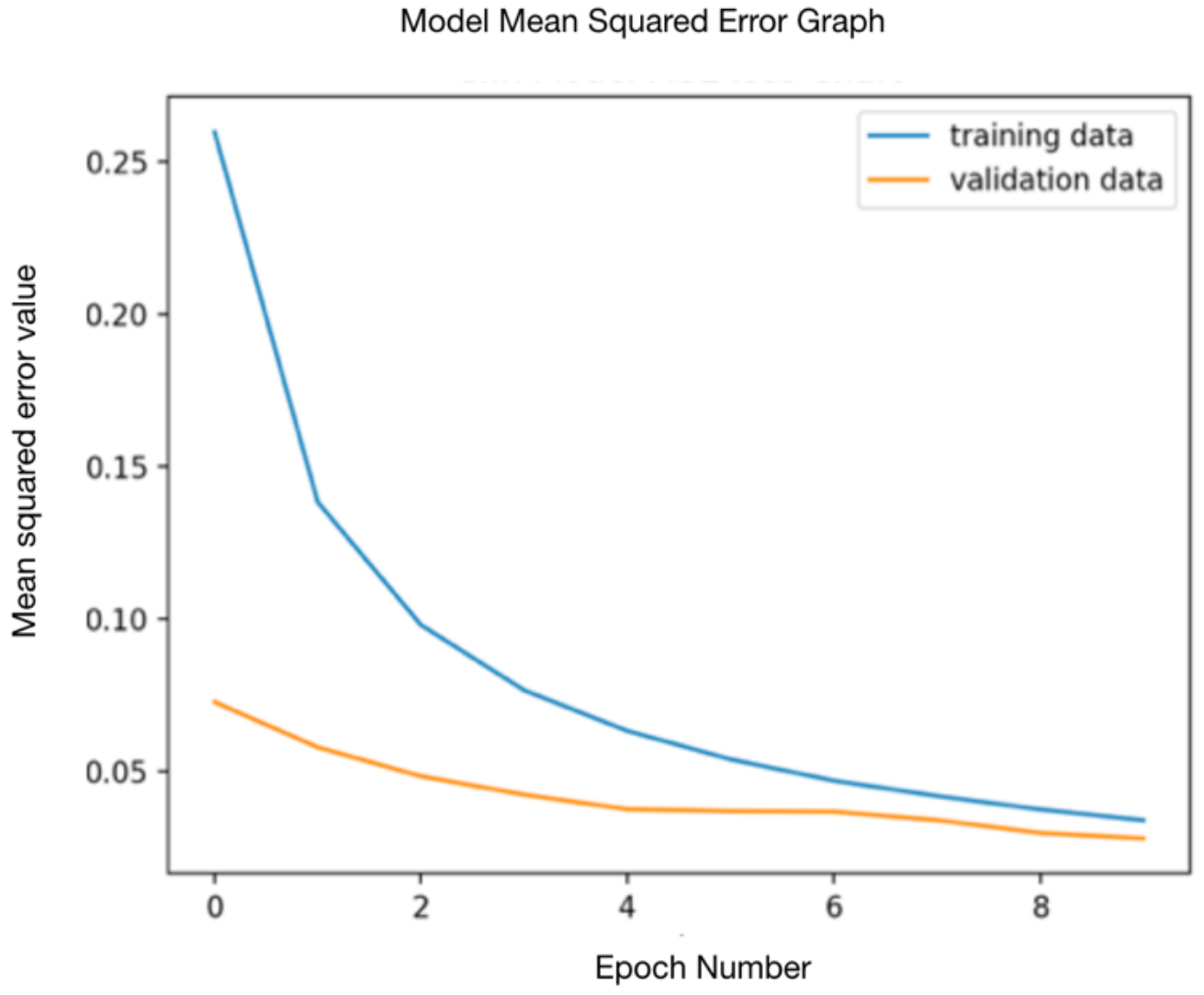

Fig. 8. Graph showing the mean squared errors of the convolutional neural network model during training. Model shows very good learning and a desired absence of overfitting.

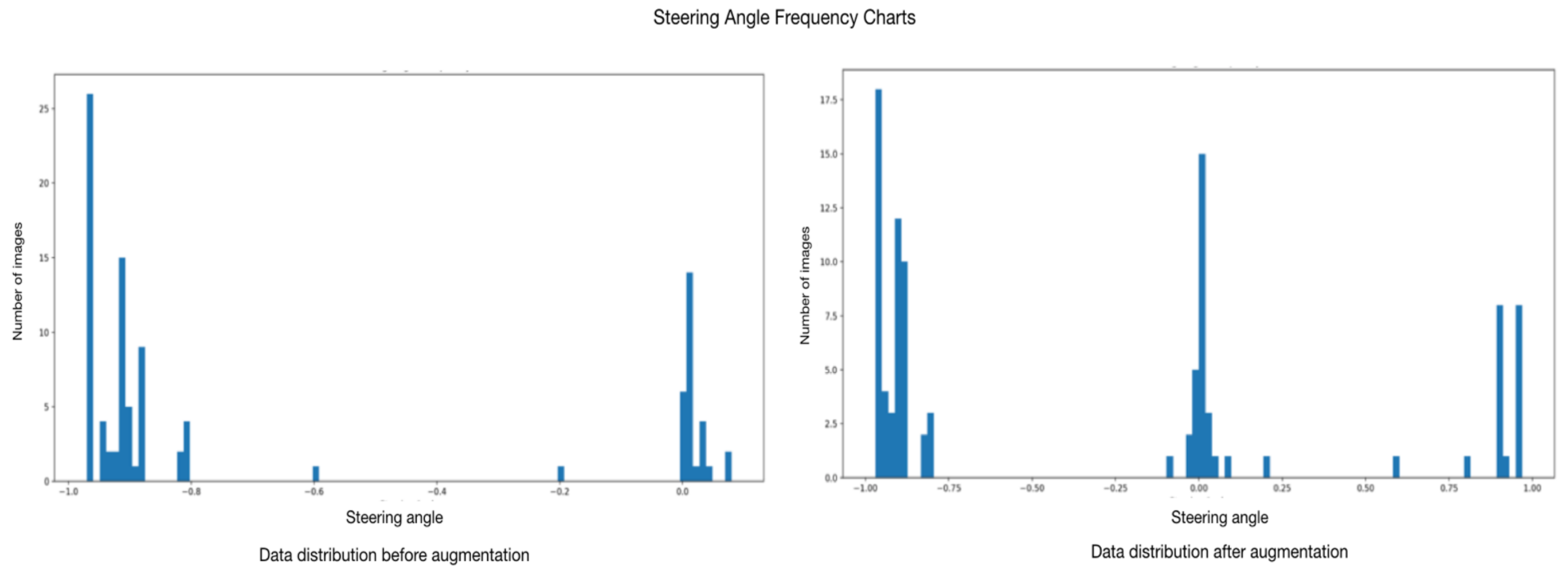

Fig. 9. Data distribution charts before and after data augmentation

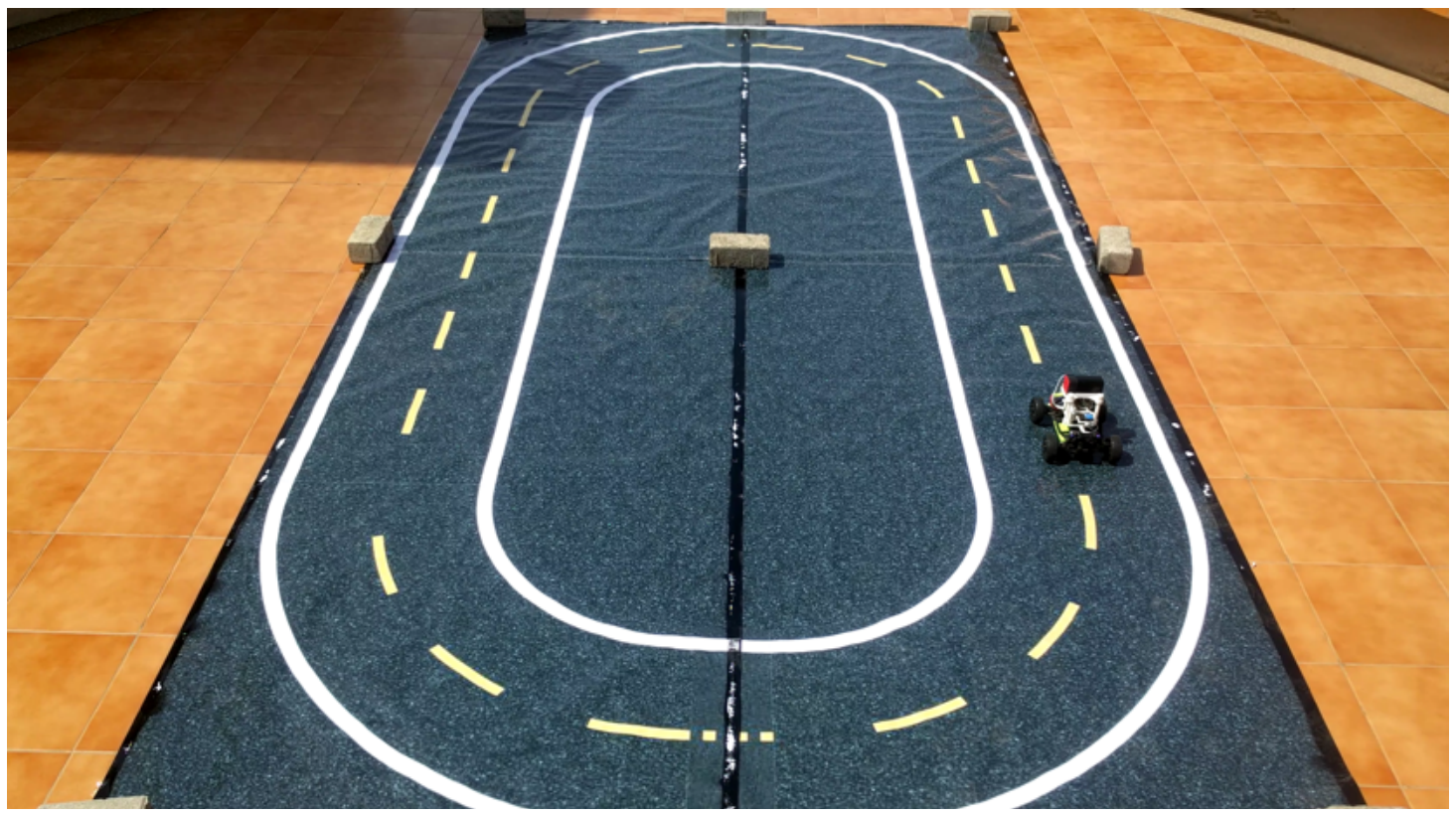

Fig. 10. RollE driving on track

## VI. Conclusion

This paper discussed the design of RollE, a low-cost proportional steering autonomous vehicle development platform. We have demonstrated with RollE that an affordable autonomous vehicle (~$250) built with hobby grade electronics can effectively be used to implement self-driving ideas using scalable technology such as machine learning. The modular nature of the software and hardware infrastructure of RollE provides an affordable skeleton that would enable further developmental work on the independent layers. RollE is designed to be a research tool, equipping self-driving researchers with a low-cost platform to test ideas and build autonomous driving technology.

While the default end-to-end machine learning model works quite well, further work can be done using the tools provided by RollE to develop and test other machine learning techniques. Future research on self-supervised learning methods would be useful to eliminate the possibility of inconsistent data collection. The physical layer would also benefit from the addition of a Global Positioning System (GPS) module, ultrasonic sensors and Light Detection and Ranging (LIDAR) sensor for precise localization, obstacle avoidance and advanced perception respectively.